\documentclass[runningheads]{llncs}
\usepackage{graphicx}
\usepackage{multirow}
\usepackage{multicol}
\usepackage{color}
\usepackage{hyperref}
\usepackage{amsmath}
\begin{document}
\title{Long-tailed multi-label classification with noisy label of thoracic diseases from chest X-ray} %\thanks{Supported by organization x.}}
\titlerunning{LTML classification of thoracic diseases from CXR}
\author{Haoran Lai\inst{1}, Qingsong Yao\inst{2}, Zhiyang He\inst{3}, Xiaodong Tao\inst{3}, S Kevin Zhou\inst{1}}

\authorrunning{Haoran Lai et al.}
\institute{University of Science and Technology of China, Hefei, China\\ \and
University of the Chinese Academy of Sciences, Beijing, China \and iFlytek, Hefei, China\\ }
% \email{skevinzhou@ustc.edu.cn} 
% \email{xdtao@iflytek.com}
%
\maketitle              % typeset the header of the contribution
\begin{abstract}
Chest X-rays (CXR) often reveal rare diseases, demanding precise diagnosis. However, current computer-aided diagnosis (CAD) methods focus on common diseases, leading to inadequate detection of rare conditions due to the absence of comprehensive datasets. To overcome this, we present a novel benchmark for long-tailed multi-label classification in CXRs, encapsulating both common and rare thoracic diseases. Our approach includes developing the ``\textbf{LTML-MIMIC-CXR}'' dataset, an augmentation of MIMIC-CXR with 26 additional rare diseases. We propose a baseline method for this classification challenge, integrating adaptive negative regularization to address negative logits' over-suppression in tail classes, and a large loss reconsideration strategy for correcting noisy labels from automated annotations. Our evaluation on LTML-MIMIC-CXR demonstrates significant advancements in rare disease detection. This work establishes a foundation for robust CAD methods, achieving a balance in identifying a spectrum of thoracic diseases in CXRs. Access to our code and dataset is provided at:~\href{https://github.com/laihaoran/LTML-MIMIC-CXR}{https://github.com/laihaoran/LTML-MIMIC-CXR}.
\end{abstract}
\section{Introduction}
 Chest X-ray (CXR) abnormalities encompass a wide variety of diseases, which can be divided into common and rare diseases. With the development of artificial intelligence, deep learning-based computer-aided diagnosis (DL-CAD) for CXR diseases has shown great potential in the diagnosis procedure of common diseases~\cite{ma2019multi,hermoza2020region,liu2022nvum}. The development of CXR DL-CAD requires a large-scale well-labeled dataset including multiple disease categories. However, existing public CXR datasets, such as NIH ChestXRay14~\cite{wang2017chestx} and MIMIC-CXR~\cite{johnson2019mimic}, only contain 13 labeled common diseases and a ``No Finding" under the setting of \underline{multi-label} classification, which greatly limits the progress of DL-CAD in CXR rare diseases diagnosis. 

 Accordingly, it is necessary to propose a large dataset that includes a large number of common and rare diseases of CXR. A previous study~\cite{RN20Arxiv} proposes a long-tailed CXR dataset by annotating extra disease labels from NIH ChestXRay14 and MIMIC-CXR to explore the potential of rare disease classification, which involves 7 extra rare diseases in the \textit{multi-class} setting. In this paper, we herein construct a new large long-tailed dataset based on MIMIC-CXR, named LTML-MIMIC-CXR, which contains 39 diseases under the \underline{multi-label} setting. First, a large number of chest diseases and their synonyms are selected from Radiology Gamuts~\cite{Budovec2014Radiographics}. Then, CheXpert~\cite{wang2017chestx} is introduced to retrieve diagnostic reports from MIMIC, which can annotate disease automatically on diagnostic reports for their paired CXR. Subsequently, we build a new large long-tailed multi-label dataset and make it public, hoping that it promotes the development of an effective DL-CAD diagnosis system.

 At the same time, several works try to improve accuracy of rare classes in the long-tailed multi-label classification task. \textit{e.g.,} Yuan \emph{et al.}~\cite{yuan2019long} adopt a two-stage training strategy, which trains a model with balanced classes first as well as a model for each long-tailed class respectively, to tackle the challenge of each long-tailed label with imbalanced distribution. Wu \emph{et al.}~\cite{wu2020distribution} propose the Distribution-Balanced Loss with class-aware enhanced resampling and the negative-tolerant regularization, which alleviates the over-suppression problem of negative logits. Guo \emph{et al.}~\cite{guo2021long} train two branches with uniform and re-balanced samplings by binary-cross-entropy-based and cross-branch-based losses, which handles the label co-occurrence by resampling. 
 
 Inspired by the above methods in computer vision~\cite{yuan2019long,wu2020distribution,guo2021long}, we develop a long-tailed multi-label classification method, %in medical image. In this paper, we aim at solving the long-tailed multi-label classification problem in the field of medical imaging analysis, and propose a 
which is deep learning-based and end-to-end. First, a novel \underline{adaptive negative regularization} is introduced to alleviate the over-suppression problem of negative logits caused by a long-tailed distribution. Second, a novel training strategy called \underline{large loss reconsideration} is applied in training stage to handle the noisy-label problem. Finally, we evaluate the effectiveness of our method in LTML-MIMIC-CXR, which achieves superior performance in ``Head" and ``Tail" classes simultaneously.

Our contributions can be summarized as follows:
\begin{itemize}
\item[$\bullet$] We establish a long-tailed multi-label CXR dataset named ``LTML-MIMIC-CXR" containing extra 26 rare diseases (39 diseases in total) under the multi-label setting, which can serve as a benchmark to aid the development of comprehensive CAD method for CXR disease diagnosis.

\item[$\bullet$] We design a novel deep learning-based method to address the challenges of long-tailed distribution and noisy label in LTML-MIMIC-CXR, which improve the total balanced accuracy from 0.5634 to 0.6484.
\end{itemize}

\section{LTML-MIMIC-CXR Dataset Construction}

\begin{figure}[t]
\includegraphics[width=\textwidth]{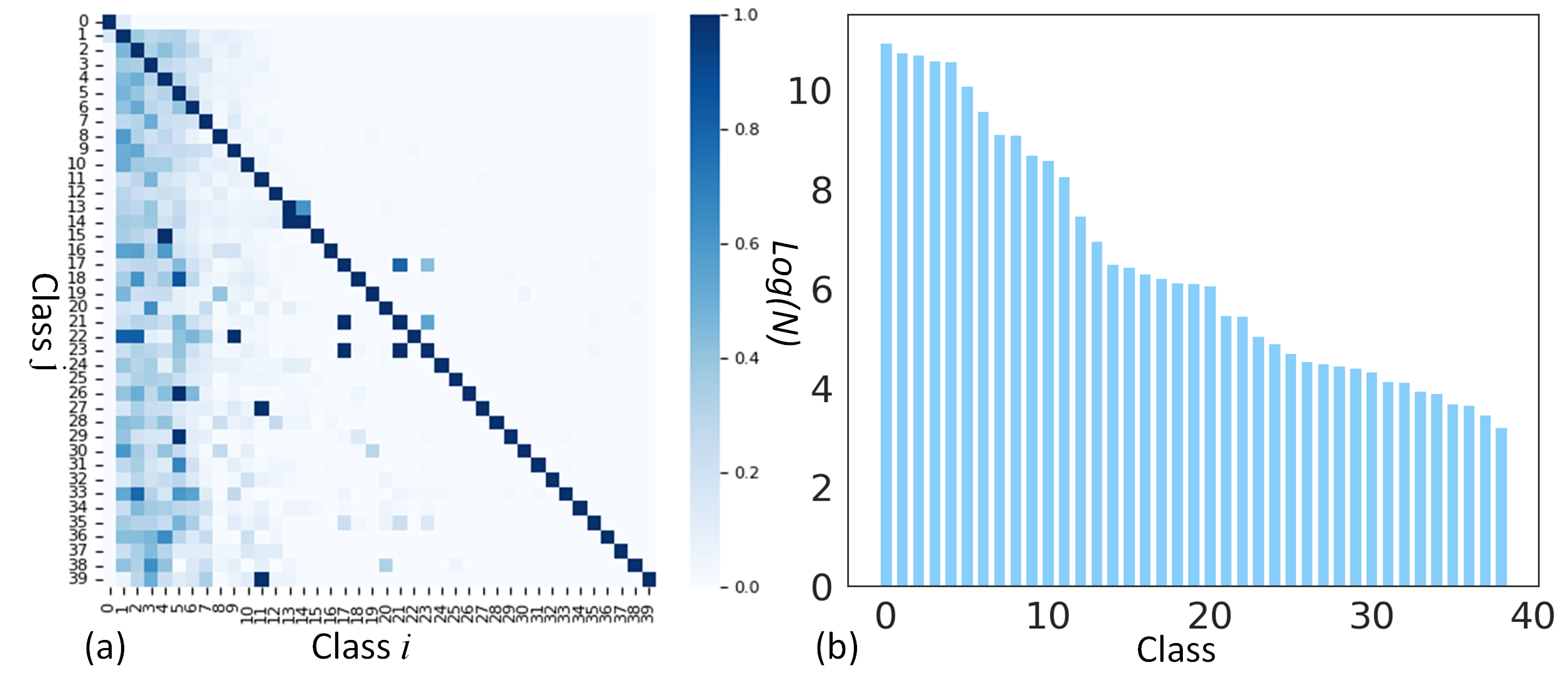}
\caption{(a) The conditional probability of class $i$ given class $j$, representing the correlation between different classes. (b) The number of samples in each class demonstrates a long-tailed distribution. To visually represent the vast differences in class frequencies, we employ $Log(N)$ on the number of each class.} \label{fig:Condition_prob}
\end{figure}

In addressing the limitations of CAD systems in diagnosing rare diseases from CXRs, we introduce the LTML-MIMIC-CXR dataset, aimed at facilitating long-tailed multi-label classification in CXR diagnosis. Building on the MIMIC-CXR dataset, we expand the disease categories from the original 13 common diseases to include an additional 26 rare diseases, sourced from Radiology Gamuts~\cite{Budovec2014Radiographics}. This expansion is complemented by re-annotating CXR images using CheXpert, excluding these common diseases. The resultant dataset comprises 39 diseases across 206,477 CXRs, divided into ``Head'', ``Medium'', and ``Tail'' classes based on disease prevalence (Fig.~\ref{fig:Condition_prob}). A stratified sampling strategy is used to divide the dataset into training, validation, and testing sets in a 7:1:2 ratio. To ensure data quality, particularly in the presence of noisy labels from automated annotation, manual verification is conducted on the testing set's labels. The LTML-MIMIC-CXR dataset, characterized by its long-tailed distribution, multi-label classification, and noisy labels, offers a significant resource for developing robust CAD systems capable of recognizing a wide spectrum of thoracic diseases.

\begin{figure}[t]
\includegraphics[width=\textwidth]{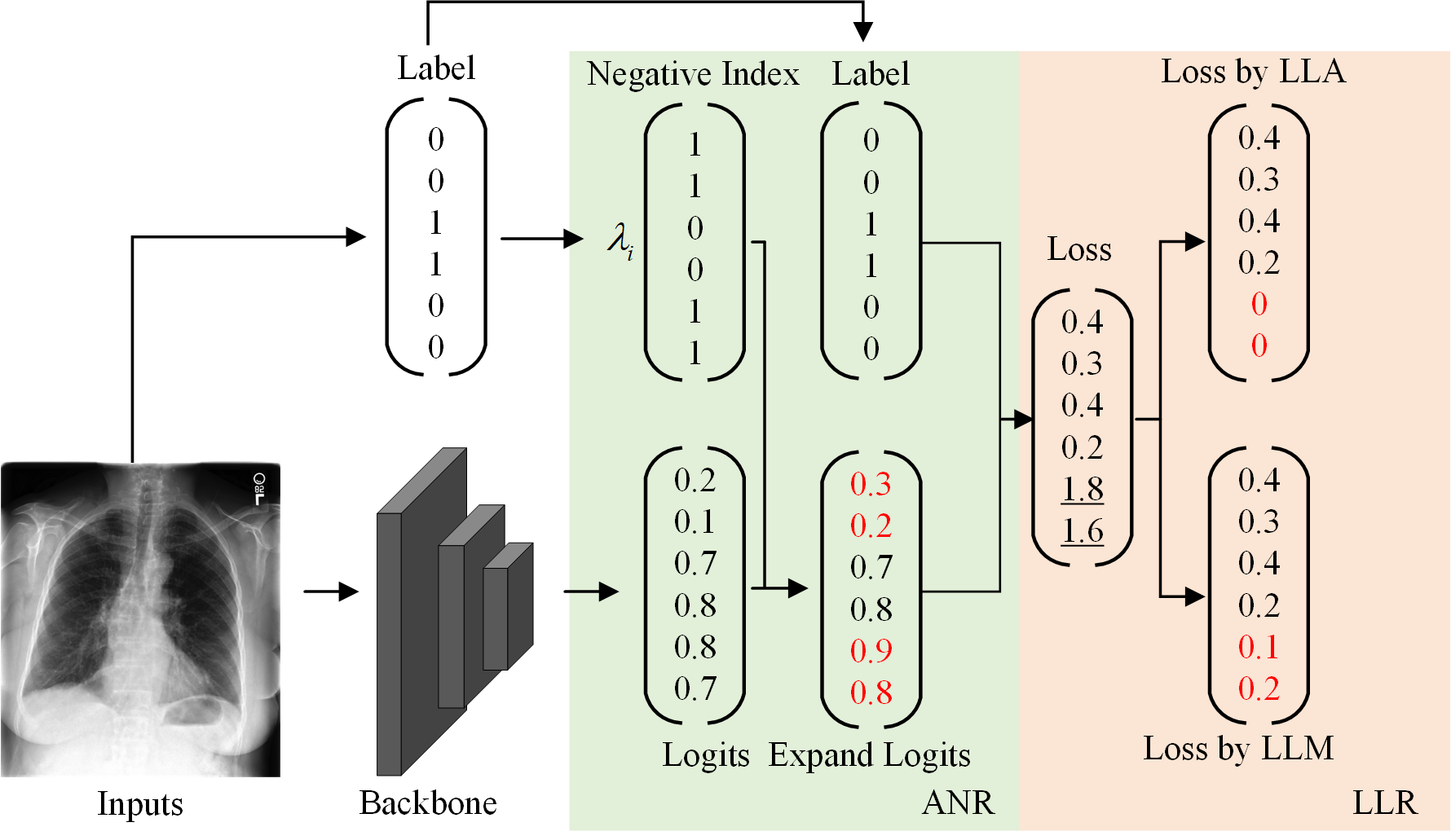}
\caption{The pipeline of our proposed method.} 
\label{fig:method}
\end{figure}

\section{Method}
As shown in Fig.~\ref{fig:method}, we propose a deep learning-based end-to-end pipeline to achieve the automatic diagnosis of CXR diseases on LTML-MIMIC-CXR. It consists of two novel components: (i) adaptive negative regularization (ANR) and (ii) large loss reconsideration(LLR).

\begin{figure}[t]
\includegraphics[width=\textwidth]{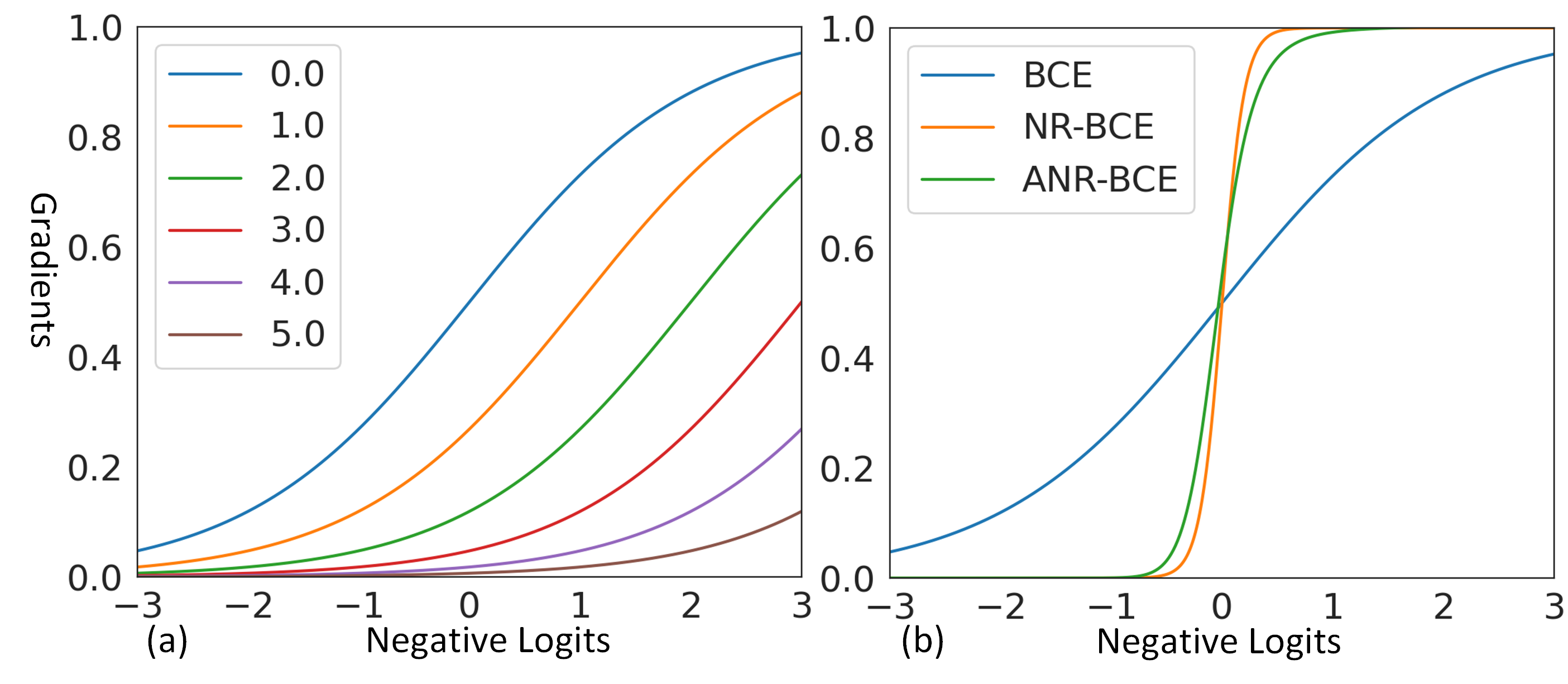}
\caption{Visualization of gradients to negative logits, including (a) the gradients of CE loss with different positive logits, and (b) the gradients of three kinds of BCE losses.} 
\label{fig:gradinet}
\end{figure}

In multi-label classification, Binary Cross-Entropy (BCE) loss with sigmoid activation is commonly used. In contrast, Cross-Entropy (CE) loss for multi-class classification typically employs softmax activation. As shown in Fig.~\ref{fig:gradinet}, the gradients of CE loss for negative logits in the presence of different positive logits can be small with a high positive logit, leading to a slower optimization process for negative logits. However, unlike softmax, the sigmoid function does not consider the correlation between different categories during computation. Wu \emph{et al.}~\cite{wu2020distribution} demonstrate that classifier outputs may overfit to tail classes in long-tailed classification, resulting in a bias towards negative classes and an over-suppression of negative logits. To address this issue, Wu \emph{et al.}~\cite{wu2020distribution} propose a linear scaling of negative logits to regulate the negative class. However, this simple linear scaling overlooks the gap between different classes.

To overcome this limitation, we propose an adaptive negative regularization (ANR), which applies a non-linear operator to adaptively scale the negative logits before computing the BCE loss. The ANR-BCE loss is formulated as:

\begin{equation}
    \mathcal{L}_{\text{positive}} = \frac{1}{C}\sum_{i=0}^{C} \left\{ y_{i} \log \left(1+e^{-u_{i}}\right)\right\};
\end{equation}

\begin{equation}
    \mathcal{L}_{\text{negative}} = \frac{1}{C}\sum_{i=0}^{C} \left\{ \frac{1}{\lambda_{i}}\left(1-y_{i}\right) \log \left(1+e^{\lambda_{i} u_{i}}\right) \right\};
\end{equation}

\begin{equation}
\label{param_u}
\lambda_{i} = 1+\beta\left(1-\sigma\left(u_{i}\right)\right),
u_{i} = p_{i}-v_{i},;
\end{equation}

\begin{equation}
    \mathcal{L}_{\text{ANR-BCE}} = \mathcal{L}_{\text{positive}} + \mathcal{L}_{\text{negative}};
\end{equation}

\noindent
where $\beta$ represents a basic scaling hyper-parameter, determining the extent of non-linear scaling applied to negative logits. The term $\lambda_{i}$ is an adaptive negative scale factor, designed to amplify the loss gradient for negative logits. Here, $C$ denotes the total number of classes. The variable $y_{i}$ signifies the class label, while $p_{i}$ corresponds to the classifier output. The function $\sigma\left(\cdot\right)$ refers to the sigmoid activation. Additionally, $v_{i}$ is introduced as a bias term for specific classes, aiming to account for model bias and formulating as:

\begin{equation}
    v_{i} = log( \frac{N}{n_{i}} - 1).
\end{equation}

\noindent
where $N$ is total number of sample and $n_{i}$ is the number of positive sample.

\subsection{Large Loss Reconsideration}
With the adoption of automatic annotation algorithms, noisy labels may arise in the LTML-MIMIC-CXR dataset. We hypothesize that while CheXpert can reliably detect positive samples, errors predominantly occur in the negative samples it identifies. To mitigate the impact of these noisy labels, we implemented a training strategy named Large Loss Reconsideration (LLR) during the training phase.

A deep neural network, in its early training stages, tends to learn from simpler samples, progressively tackling harder samples in the middle and later stages~\cite{Kim_2022_CVPR}. In scenarios with noisy labels, these hard samples often represent incorrectly labeled instances, resulting in substantial losses that can impede model optimization. Given that a large loss is indicative of a potential mislabeling, it is prudent to reassess these losses prior to gradient computation. Furthermore, aligning with our assumption that incorrect labeling predominantly affects negative samples in LTML-MIMIC-CXR, we propose two approaches to reevaluate large losses: (1) Large Loss Abandoning (LLA), which involves disregarding certain losses, and (2) Large Loss Modifying (LLM), which entails reclassifying select negative samples as positive before gradient calculation. Specifically, during each training iteration, we identify the top $k$ largest losses among the negative samples from the batch loss. These are then subjected to LLR before the gradient computation. As the training progresses and the model becomes more adept at identifying inaccurately labeled samples, the value of $k$ for reconsideration is adjusted upwards accordingly.

\section{Experiments}

\subsection{Implementation Details}
 We utilize a ResNet50 pre-trained on ImageNet as our feature extractor backbone. Image preprocessing includes $z$-score normalization and resizing to $224 \times 224$. To mitigate the long-tailed distribution, we apply class-aware enhanced resampling and data augmentation during training. Focal loss is adopted as the loss function, with the hyper-parameter $\beta$ empirically set to 10. Optimization is carried out using SGD with 0.9 momentum and $1 \times 10^{-4}$ weight decay, a learning rate of 0.02, a batch size of 32. Experiments are conducted on PyTorch, using an RTX 3090 GPU.

\subsection{Evaluation Metrics}
Balanced accuracy (BACC)~\cite{brodersen2010balanced} and area under the curve (AUC) are introduced to evaluate the performance of all classes. Moreover, we report the BACC and AUC of each subset, namely ``Head," ``Medium," and ``Tail," to observe how the techniques affect them.

\begin{table}[t]
\caption{Ablation experiment of our proposed methods on testing set.}
\label{tab:ablationstudy}
\begin{center}
\begin{tabular}{lcccccccc}
\hline
~ &  \multicolumn{4}{c}{BACC} &  \multicolumn{4}{c}{AUC} \\
 \cline{2-5} \cline{6-9}
Methods &  Total & Head & Medium & Tail & Total & Head & Medium & Tail\\
\hline

NR &  56.34 & 59.92 & 53.16 & 56.23 &  72.18 & 76.06 & 68.94 & 72.00\\

ANR &  58.41 & 65.04 & 51.09 & 58.60 &  72.49 & 75.92 & \textbf{69.96} & 72.24\\

ANR-LLA &  \textbf{64.84} & 64.85 & \textbf{59.91} &\textbf{66.21} &  \textbf{74.56} & \textbf{76.09} & 69.39 & 75.58\\

ANR-LLM &  64.14 & \textbf{66.53} & 57.93 & 65.21 &  74.50 & 74.85 & 66.00 & \textbf{76.78}\\

\hline
\end{tabular}
\end{center}
\end{table}

\subsection{Results and Discussion}

\subsubsection{Ablation Study}
\begin{figure}[t]
\includegraphics[width=0.9\textwidth]{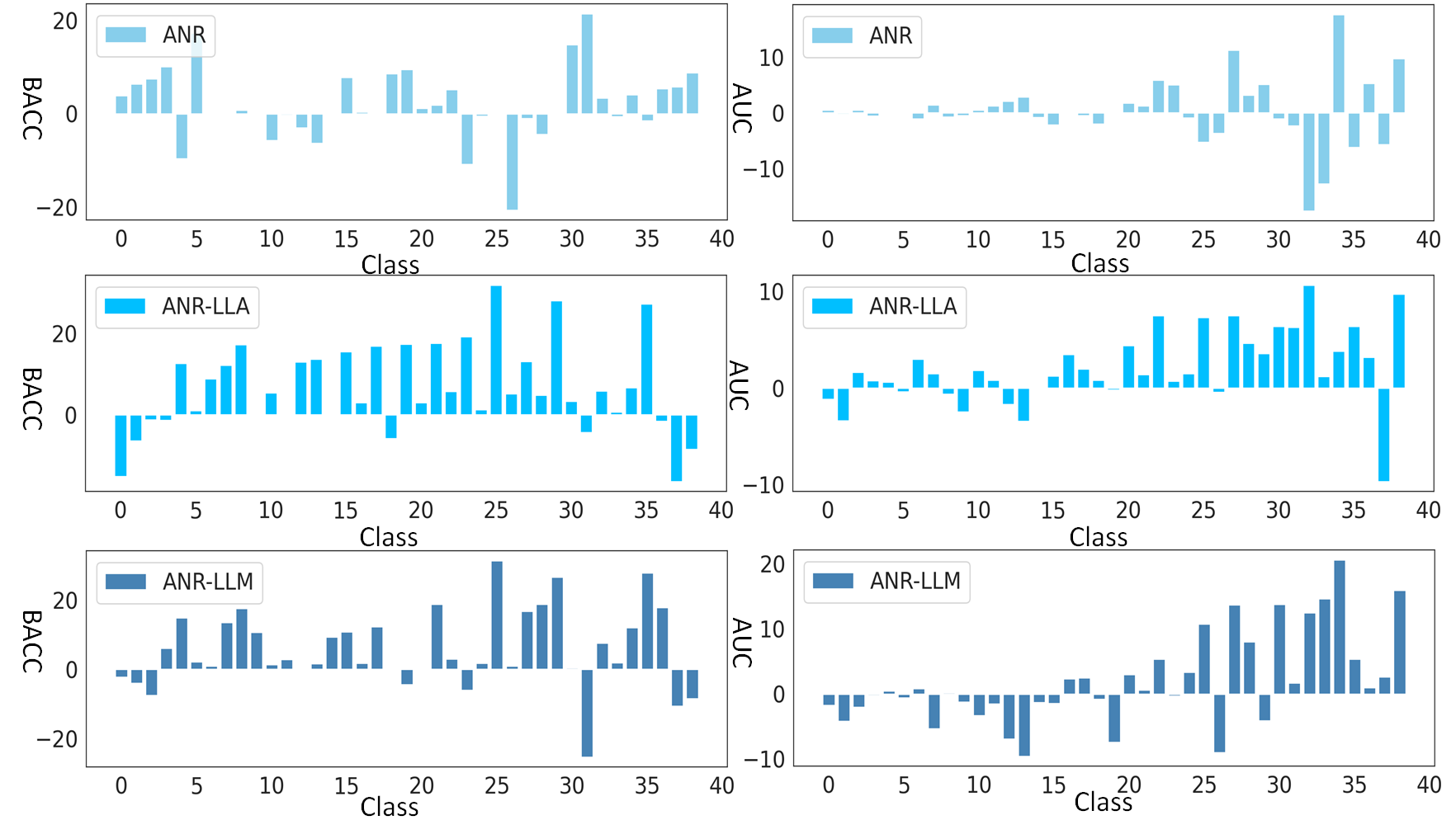}
\caption{The per-class BACC and AUC increments between every two steps is performed to evaluate our proposed method, which adopts NR as baseline.} 
\label{fig:class_cmp}
\end{figure}

To explore the effectiveness of each module, the ablation study is implemented by using negative regularization (NR) with linear scaling as baseline and adding the proposed modules step by step. First, as shown in Table~\ref{tab:ablationstudy}, compared with linear scaling of negative logits, our proposed adaptive negative scaling achieves increments in total BACC and AUC by 2.07 and 0.31, respectively. Moreover, as shown in Fig.~\ref{fig:class_cmp} (the 1st row), compared with NR, there is a significant improvement in both head and tail classes in ANR, which may contribute to adaptive learning. Different over-suppression of different negative logits can be expanded adaptively by ANR, thus achieving well performance.

Second, considering noisy label, LLR module is introduced to alleviate the harm of noisy label, which includes two strategies: LLR and LLM. As shown in Table~\ref{tab:compareMethod} and Fig.~\ref{fig:class_cmp} (the 2nd and 3rd rows), we find that LLR module led to a significant performance improvement, especially in BACC. Moreover, the performance improvement in LLA is greater than that in LLM. Since LLM may introduce new label errors during the process of correcting labels for large losses, LLA simply ignores negative logits with large losses, which avoids introducing new label errors and achieves better overall performance than LLM. Therefore, our proposed LLR module can be used to handle noisy label effectively on LTML-MIMIC-CXR. 

\begin{table}[t]
\caption{Comparison of different methods on testing set. }
\label{tab:compareMethod}
\begin{center}
\begin{tabular}{lcccccccc}
\hline
~ &  \multicolumn{4}{c}{BACC} &  \multicolumn{4}{c}{AUC} \\
 \cline{2-5} \cline{6-9}
Methods &  Total & Head & Medium & Tail & Total & Head & Medium & Tail\\
\hline

BCE loss &  54.18 & 65.49 & 53.73 & 51.15 &  71.12 & \textbf{76.59} & \textbf{69.82} & 69.95\\

Focal loss~\cite{lin2017focal} &  54.19 & 65.16 & 53.47 & 51.31 &  71.54 & 76.29 & 69.54 & 70.78\\

DB loss~\cite{wu2020distribution} &  56.34 & 59.92 & 53.16 & 56.23 &  72.18 & 76.06 & 68.94 & 72.00\\

NVUM~\cite{liu2022nvum} & 55.51 & 66.40 & 53.05 & 53.15 & 73.06 & 76.54 & 69.56 & 73.07 \\
\hline
ANR-LLA &  \textbf{64.84} & 64.85 & \textbf{59.91} &\textbf{66.21} &  \textbf{74.56} & 76.09 & 69.39 & 75.58\\

ANR-LLM &  64.14 & \textbf{66.53} & 57.93 & 65.21 &  74.50 & 74.85 & 66.00 & \textbf{76.78}\\
\hline
\end{tabular}
\end{center}
\end{table}

\subsubsection{Comparison Methods}
The comparative methods performed in this study can be introduced as follow: (1) BCE loss; (2) Focal loss~\cite{lin2017focal}; (3) Distribution Balanced (DB) loss~\cite{wu2020distribution}, which introduced negative-tolerant regularization to handle long-tailed problem in multi-label classification; (4) Non-volatile Unbiased Memory (NVUM)~\cite{liu2022nvum}, which applied unbiased memory to regularize the classifier logits for noisy label on CXR dataset. For fair comparison, class-aware enhanced resampling~\cite{wu2020distribution} is applied in all experiments to alleviate the influence of long-tailed distribution. As shown in Table~\ref{tab:compareMethod}, BCE loss achieves best performance in AUC of ``Head" and ``Medium" classes but worst in ``Tail" classes, possibly due to the larger number of samples in ``Head" and ``Medium" classes, which makes it easier for the BCE loss to learn discriminative features. Compared with existing methods, our method achieves the best performance in total BACC and AUC, especially in tail class, which illustrates the advantage of our method on long-tailed multi-label classification with noisy label.

\section{Conclusion}
The development of CXR DL-CAD is of great significance for disease diagnosis. However, there is still limited research on rare diseases in CXR. In this paper, we have constructed a new CXR long-tail multi-label dataset, named LTML-MIMIC-CXR. Furthermore, we have proposed an end-to-end pipeline to effectively address the challenges presented in LTML-MIMIC-CXR. First, we introduce the ANR module to alleviate the over-suppression of negative logits by long-tailed distribution. Second, we incorporate LLR module to handle noisy label by automatic annotation. The experimental results demonstrate the effectiveness of our proposed method, which can serve as a baseline for future research on LTML-MIMIC-CXR.

\bibliographystyle{splncs04}
\bibliography{mybibliography}

\end{document}